%% file: templateArxiv.tex
\documentclass{article}

\usepackage{PRIMEarxiv}
\usepackage{tabularx}
\usepackage[utf8]{inputenc} 
\usepackage[T1]{fontenc}    
\usepackage{hyperref}       
\usepackage{url}            
\usepackage{booktabs}       
\usepackage{amsfonts}       
\usepackage{nicefrac}       
\usepackage{microtype}      
\usepackage{lipsum}
\usepackage{fancyhdr}       
\usepackage{graphicx}       
\graphicspath{{media/}}     

\title{Translating Inference-Time Control to Radiology Vision-Language Models: Activation Steering for Pneumonia Classification on Chest X-rays}

\author{
  Eduardo Moreno Judice de Mattos Farina, MD \\
  Universidade Federal de São Paulo (UNIFESP) \\
  Hospital Israelita Albert Einstein \\
  São Paulo, Brazil \\
  \And
  Mateus A. Esmeraldo, MD \\
  Department of Radiology \\
  Stanford University School of Medicine \\
  Stanford, CA, USA \\
  \And
  Felipe Akio Matsuoka \\
  Universidade Federal de São Paulo (UNIFESP) \\
  DASA \\
  São Paulo, Brazil \\
  \And
  Paulo Eduardo de Aguiar Kuriki, MD \\
  University of Texas Southwestern Medical Center (UTSW) \\
  Dallas, TX, USA \\
  \And
  Felipe Campos Kitamura, MD, PhD \\
  Universidade Federal de São Paulo (UNIFESP) \\
  Eden \\
  São Paulo, Brazil / USA \\
}

\begin{document}
\maketitle

\begin{abstract}
Inference-time engineering can alter model behavior without fine-tuning. However, its utility for improving diagnostic performance in medical vision-language models (VLMs) remains unclear. We aim to evaluate whether Contrastive Activation Addition (CAA) can improve pneumonia classification in chest radiograph VLMs without updating model weights. Three frozen chest radiograph VLMs (MedGemma-4B-IT, NV-Reason-CXR-3B, and CheXOne-3B) were evaluated on the public Kermany pneumonia test set. Classification was based on the logits of the tokens \textit{Yes} and \textit{No} under a binary prompt. Steering vectors included a 30-pair answer-bias control, a 30-pair pneumonia text contrast, and an image-conditioned contrast derived from 30 pneumonia and 30 normal development images. A deterministic 200-image development set was used for layer and scale selection (100 images) and threshold calibration (100 images). Performance was assessed using ROC-AUC, PR-AUC, F1 score, threshold analyses, reverse-vector controls, random-vector controls, and conditional bootstrap confidence intervals. Fixed-threshold F1 improvements were frequently observed but did not consistently indicate improved diagnostic performance. For MedGemma-4B-IT. NV-Reason-CXR-3B showed the strongest benefit: calibrated F1 improved from 0.7692 in the zero-shot setting to 0.8619 with pneumonia-text steering and to 0.8727 with image-conditioned steering. For CheXOne-3B, pneumonia-text steering increased calibrated F1 from 0.8528 to 0.8666, although the confidence interval crossed zero. On this public pneumonia benchmark, CAA substantially altered prediction score distributions and operating characteristics without fine-tuning. Meaningful performance gains were observed in one of three evaluated VLMs, suggesting that activation steering may serve as a lightweight approach for adapting medical VLM behavior.
\end{abstract}


\newpage

\section{Introduction}
Medical VLMs are increasingly evaluated for chest radiograph interpretation, but their behavior can depend strongly on prompt format and decision thresholds. This raises a practical question for medical AI: can a model's pneumonia classification performance be improved at inference time without fine-tuning the model weights?

Representation Engineering studies directions in activation space that can be used to inspect or influence model behavior\cite{zou2025representationengineeringtopdownapproach}. Contrastive Activation Addition constructs a steering vector from positive and negative examples and adds the vector to hidden states during inference\cite{panickssery2024steeringllama2contrastive}. Because no weights are updated, CAA is attractive as a lightweight intervention: it can be applied after training, swept across layers and scales, and combined with threshold calibration to tune operating behavior.

The same convenience creates an evaluation problem. If steering moves many examples across an arbitrary threshold, the resulting fixed-threshold F1 may look like an improved diagnosis. A stronger claim requires comparison with calibrated thresholds and control vectors.

This study evaluates CAA for CXR pneumonia classification as a form of inference-time engineering. We use "inference-time engineering" to denote CAA-based representation steering and downstream threshold calibration applied without weight updates. The aim is simple: to test whether steering improves pneumonia-classification behavior and whether the apparent gains remain after calibration and control experiments.

\section{Materials and Methods}
\begin{figure}[htbp]
    \centering
    \includegraphics[width=\textwidth]{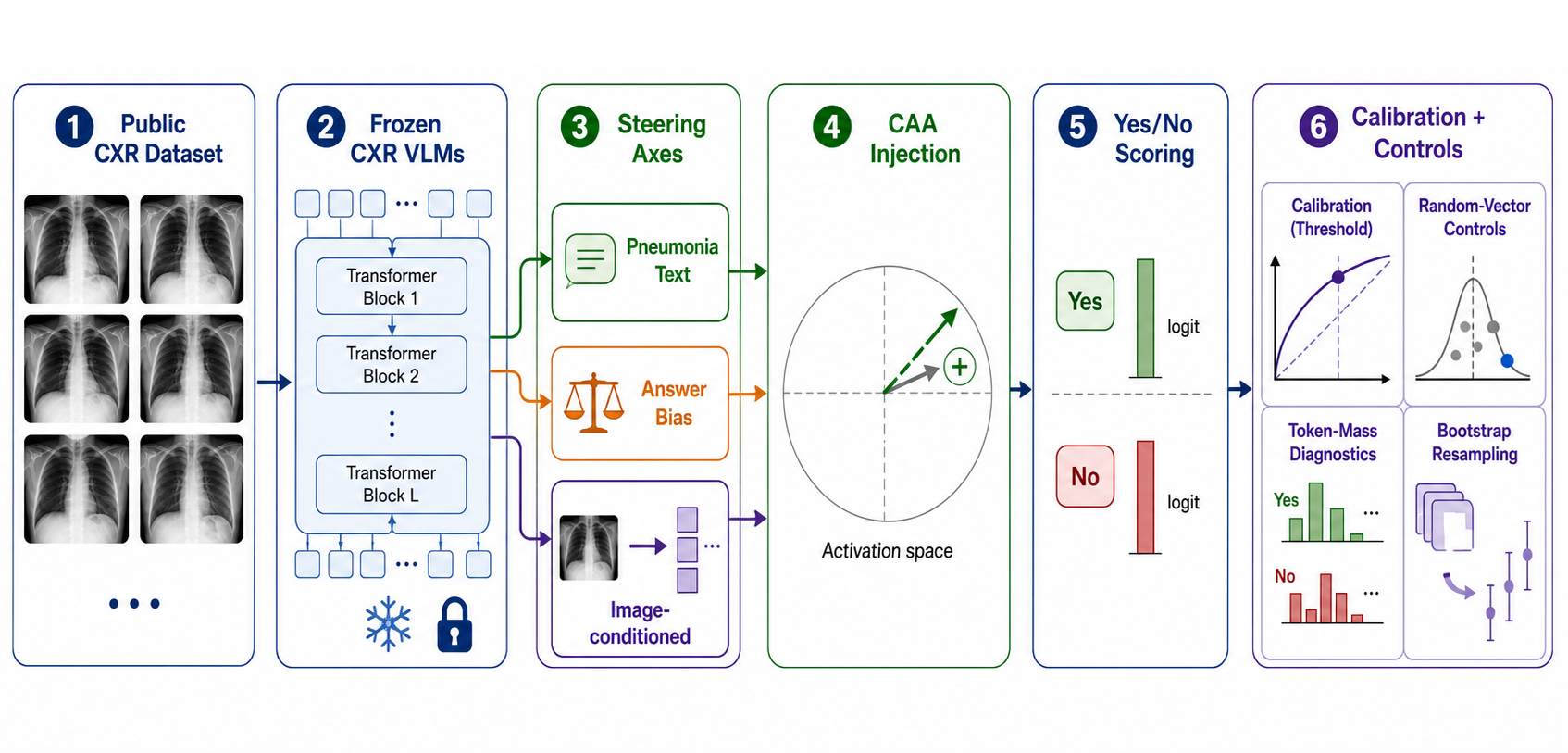}
    \caption{Methods overview. Frozen chest X-ray VLMs were evaluated with three inference-time steering axes, forced-choice Yes/No scoring, threshold calibration, and control analyses.}
    \label{fig:methods_overview}
\end{figure}

\subsection{Study Design and Dataset}
This was a retrospective experimental benchmark using public de-identified data. A general overview of the study is demonstrated in Figure ~\ref{fig:methods_overview}.
The dataset was hf-vision/chest-xray-pneumonia, a public Hugging Face mirror of the Kermany pediatric chest radiograph pneumonia dataset\cite{Kermany2018}. The complete test split was used for final evaluation and contained 624 radiographs: 390 labeled pneumonia and 234 labeled normal. The local public mirror did not include age, sex, scanner, institution, acquisition date, or other demographic and acquisition metadata. Dataset labels served as image-level ground truth.
No test-set examples were used for steering-vector construction, layer selection, scale selection, or threshold selection. Development examples were sampled from the training split using seed 42 and offset 500. The deterministic 200-image development set was split into 100 images for layer/scale selection and 100 images for threshold selection.

\subsection{Models}
Three frozen CXR VLMs were evaluated: MedGemma-4B-IT (google/medgemma-1.5-4b-it), NV-Reason-CXR-3B (nvidia/NV-Reason-CXR-3B), and CheXOne-3B (StanfordAIMI/CheXOne)\cite{myronenko2025reasoningvisuallanguagemodel, zhang2026reasoningenabledvisionlanguagefoundationmodel, sellergren2026medgemma15technicalreport}.  No model weights were fine-tuned. 

\subsection{Pneumonia Classification Prompt and Scoring}
Each model received a chest radiograph and the same binary pneumonia question: "You are an expert radiologist. Look at this chest X-ray carefully. Does this patient have pneumonia? Answer with only Yes or No." Scoring did not parse the generated free text. Instead, Yes and No were verified as single-token labels for each tokenizer, and the model was scored from the next-token logits at the last non-padding token of the formatted prompt, corresponding to the answer position.
Pneumonia was predicted when this score exceeded the evaluated threshold. This constrained score is useful for controlled comparison, but it is not automatically a calibrated probability. Therefore, the benchmark also stored full-vocabulary P(Yes), P(No), P(Yes)+P(No) mass, Yes and No token ranks, Yes-minus-No logit margins, duplicate-score counts, ROC-AUC, and PR-AUC. In low-mass regimes, ROC-AUC and PR-AUC were interpreted as rank metrics over the Yes-minus-No margin rather than calibrated confidence.
\subsection{Steering Axes}
The steering axes are described in Table ~\ref{tab:steering_axes}.

\input{table1}

\subsection{Steering Vector Construction and Injection}
For each supported axis and candidate layer, hidden activations were collected separately for the positive and negative sides of the contrast. The steering vector was computed as the mean positive activation minus the mean negative activation at the last non-padding token of the formatted prompt, then normalized before inference-time injection. 
During evaluation, model weights remained frozen. The selected unit vector was added to the selected language-decoder transformer block output with the selected scalar multiplier. For hybrid VLMs, steering was applied to the language-decoder transformer layers, not to the vision encoder or projection modules. The vector was constructed from the last non-padding-token activations, but injection was broadcast across token positions at the selected decoder block output during the steered forward pass.
\subsection{Layer and Scale Selection}

Candidate steering layers were defined as relative depths within the decoder stack, corresponding to approximately 35\%, 45\%, 55\%, 65\%, and 75\% of model depth, indexed from the output layer backward. Candidate steering scales were 1, 2, 5, 10, 20, 50, 100, 200, and 500. For each model--axis pair, all layer-scale combinations were evaluated on the 100-image selection subset. The optimal configuration was defined as the combination yielding the highest F1 score, with prespecified tie-breaking by sensitivity, accuracy, and smaller absolute scale.

Selected steering scales varied substantially across models, reflecting differences in residual-stream magnitudes and internal representations. Accordingly, the reported scales should be interpreted as model-specific hyperparameters selected for reproducibility rather than as generally transferable settings. Future studies may benefit from normalization strategies that account for activation norms when determining steering coefficients.

\subsection{Threshold Calibration}
Zero-shot and steered thresholds were not shared. Each condition selected its own threshold on the independent 100-image threshold split, after layer and scale selection had already been fixed. Candidate thresholds were derived from the observed development scores, and the selected threshold maximized F1 on the threshold split. Final calibrated metrics were then computed once on the untouched test split using development-selected thresholds. Fixed-threshold metrics at 0.5 were retained as sensitivity analyses. 
Conditional bootstrap CIs held the selected thresholds fixed. Nested threshold-selection bootstrap CIs resampled the threshold-development split and paired test examples and reselected thresholds inside each bootstrap replicate. This nested procedure addressed only threshold-selection variability; vector construction and layer/scale selection were not re-nested.

\subsection{Controls}
Reverse-vector controls used the selected vector with a negative scale at the selected layer. Random-vector controls used 20 deterministic random unit-vector seeds per model and axis at the selected layer and absolute scale. Answer-bias steering served as a semantic control for direct Yes/No response-prior movement. Image-conditioned steering served as a complementary axis in which positive and negative text were identical, and the contrast came from the image label.

\section{Results}

\begin{figure}[htbp]
    \centering
    \includegraphics[width=\textwidth]{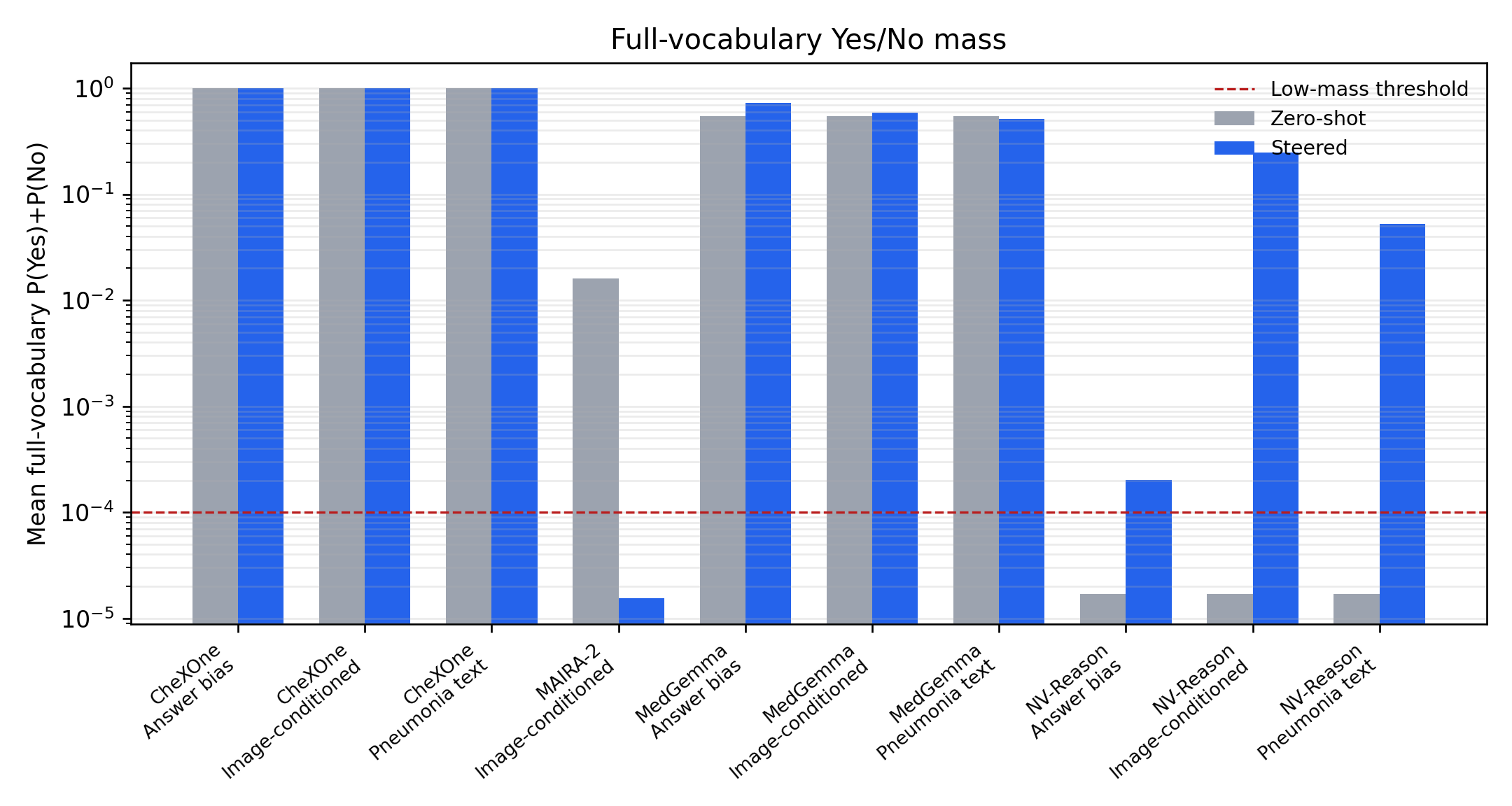}
    \caption{Mean full-vocabulary Yes/No probability mass. Mass is $P_{\mathrm{vocab}}(\mathrm{Yes}) + P_{\mathrm{vocab}}(\mathrm{No})$; low-mass regimes should be interpreted as margin-ranking diagnostics rather than calibrated clinical confidence.}
    \label{fig:probability_mass}
\end{figure}

\subsection{Forced-Choice Validity Varied Substantially}
All models supported single-token Yes and No labels. However, the full-vocabulary probability mass assigned to these two answer tokens varied substantially across models and steering conditions. 
CheXOne-3B assigned nearly all first-token probability mass to Yes and No in the zero-shot setting, making constrained Yes/No scoring relatively direct. By contrast, NV-Reason-CXR-3B assigned very low zero-shot probability mass to the Yes/No pair, so its constrained score primarily reflected the relative ranking of two low-  probability tokens rather than a naturally concentrated binary answer distribution. Figure ~\ref{fig:probability_mass} shows how steering changed this full-vocabulary answer-token mass across conditions.
This diagnosis affected interpretation. Low Yes/No mass does not make a result invalid, but it changes what can be claimed from it: the analysis should be interpreted as a constrained-scoring sensitivity analysis, not as evidence that the model would naturally answer the clinical question with calibrated binary confidence.

\begin{figure}[htbp]
    \centering
    \includegraphics[width=\textwidth]{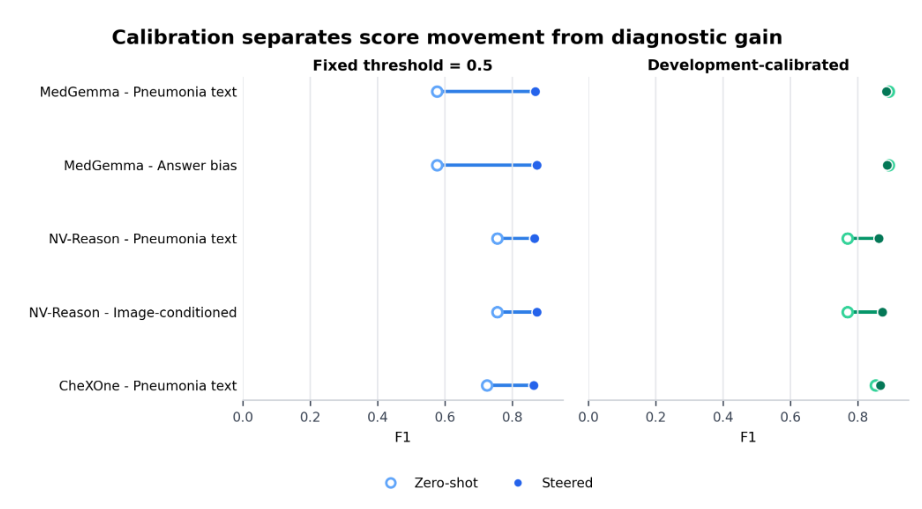}
    \caption{Line plots showing zero-shot and steered F1 scores for selected model--axis pairs at a fixed threshold of 0.5 (blue) and after threshold selection (green) on the independent development-threshold split. Open circles indicate zero-shot performance, whereas filled circles indicate steered performance.}
    \label{fig:threshold_calibration}
\end{figure}

\subsection{Fixed-Threshold Gains Were Large and Misleading}
At threshold 0.5, steering often looked impressive. MedGemma-4B-IT fixed-threshold F1 increased from 0.5761 zero-shot to  0.8681 with pneumonia-text steering, 0.8725 with answer-bias steering, and 0.8587 with image-conditioned steering.. The answer-bias result is the key control: a vector built from affirmative versus negative answer phrasing produced a gain comparable with disease axes. Figure ~r\ref{fig:threshold_calibration} contrasts these fixed-threshold results with independently calibrated operating points.

\begin{figure}[htbp]
    \centering
    \includegraphics[width=\textwidth]{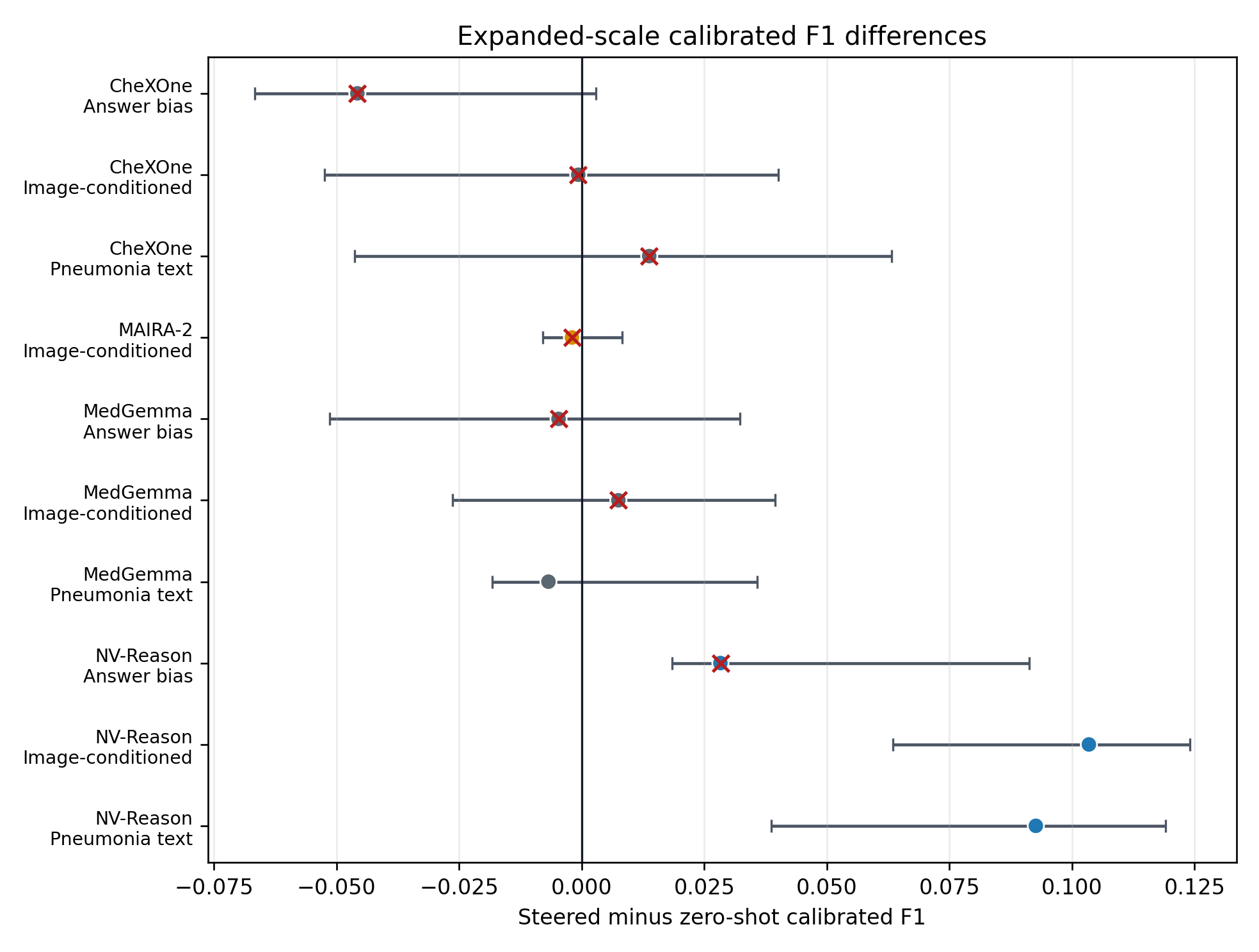}
    \caption{Steered minus zero-shot calibrated F1 differences after the expanded-scale run. Error bars show nested threshold-selection bootstrap 95\% confidence intervals. X markers indicate fixed-threshold overlap with the random-vector range.}
    \label{fig:f1_differences}
\end{figure}

\subsection{Threshold Calibration Changed the Main Result}
After independent threshold calibration, the main result shifted from broad fixed-threshold gains to a more selective pattern of model- and axis-specific effects. MedGemma-4B-IT was the clearest example: calibrated zero-  shot performance was already high, and none of the steered MedGemma conditions showed a robust calibrated advantage over zero-shot. Thus, the large fixed-threshold gains in MedGemma were best explained by score recalibration rather than disease-specific improvement.

The calibrated F1 differences are shown in Figure ~\ref{fig:f1_differences}. Only the NV-Reason-CXR-3B pneumonia-text and image- conditioned axes had nested 95\% CIs that excluded zero in the positive direction. This made NV-Reason the strongest positive disease-axis signal. The answer-bias condition also shifted calibrated performance, but because this axis directly targets the Yes/No response channel rather than pneumonia evidence, it was treated as a control rather than evidence of disease-specific steering.

CheXOne-3B showed a different pattern. Pneumonia-text steering produced a small calibrated F1 increase, but its nested CI crossed zero. Image-conditioned and answer-bias steering also failed to show robust, calibrated improvement. These results indicate that CheXOne steering effects were threshold-sensitive and not sufficient to support a robust diagnostic-performance claim.

Overall, threshold calibration narrowed the interpretation: the main positive finding was confined to NV-Reason disease-axis steering, while MedGemma and CheXOne primarily illustrated why fixed-threshold gains require calibration and controls. Full cell-level metrics, PR-AUC values, Yes/No mass, and nested CIs are reported in Table ~\ref{tab:expanded_performance}.

\input{table2}

\subsection{Random-Vector Controls Exposed Nonspecific Effects}
\begin{figure}[htbp]
    \centering
    \includegraphics[width=\textwidth]{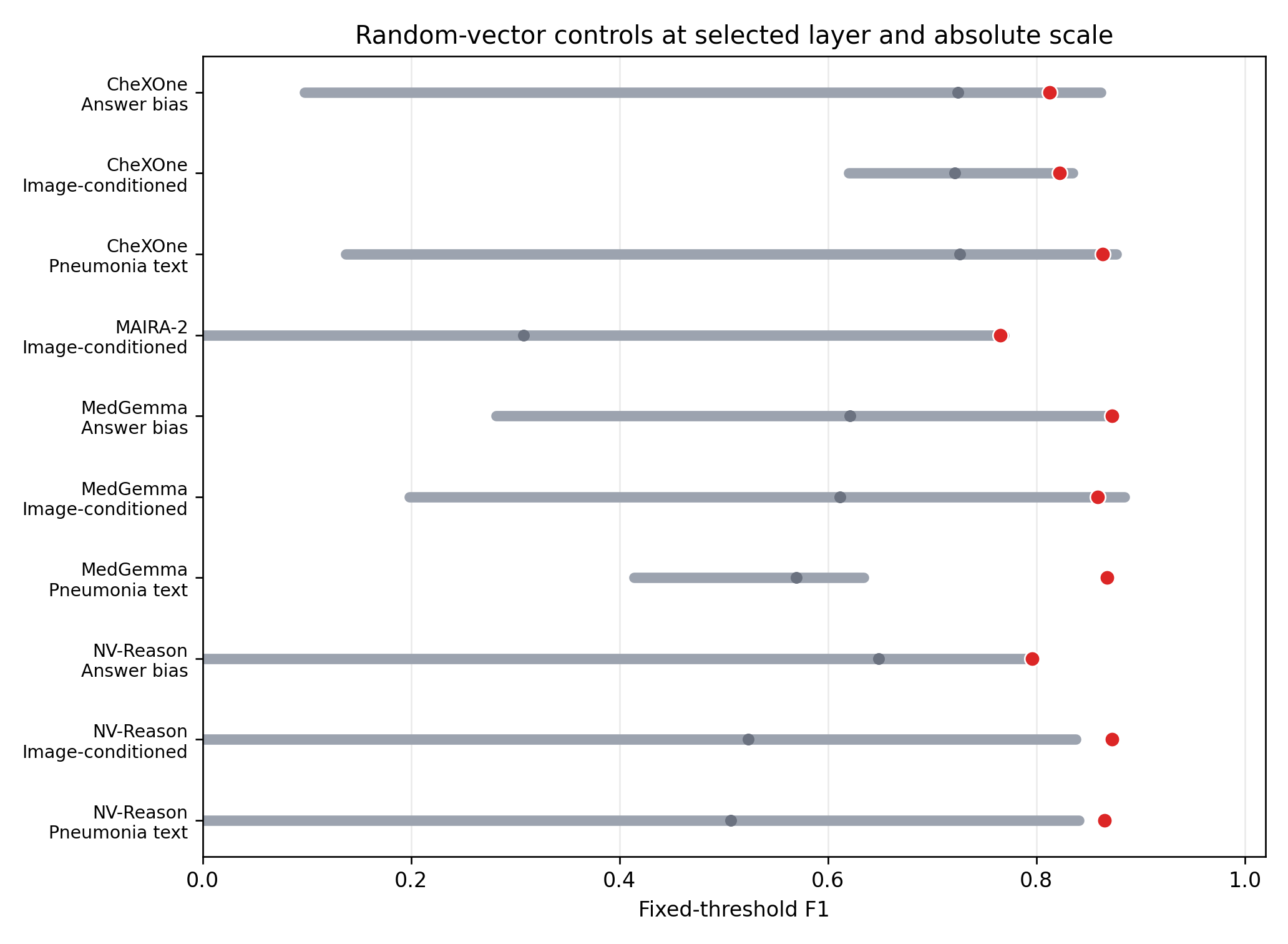}
    \caption{Random-vector controls. Gray ranges show fixed-threshold F1 across 20 deterministic random vectors; red dots indicate selected steering conditions..}
    \label{fig:random_vectors}
\end{figure}

Random-vector controls showed that activation injection itself could produce large fixed-threshold F1 shifts (Figure ~\ref{fig:random_vectors}). In several conditions, the selected steering vector did not clearly separate from the distribution of  20 random unit vectors applied at the same layer and absolute scale. This was most evident for CheXOne, where the selected pneumonia-text, answer-bias, and image-conditioned vectors all fell within or near the random-vector range. Similar overlap or near-overlap was also seen for MedGemma answer-bias and image-conditioned steering.  
These controls limit the interpretation of high fixed-threshold F1 values. A large fixed-threshold gain was not treated as disease-specific if random directions could produce a comparable operating-point shift. The stronger evidence came from conditions that combined calibrated F1 improvement, positive nested threshold-selection CIs,  adequate Yes/No token mass, and a plausible disease-axis interpretation.
MedGemma pneumonia-text steering illustrates the distinction between these criteria. This vector cleared the random-vector control, indicating that its fixed-threshold score movement was not simply interchangeable with arbitrary activation injection. However, after threshold calibration, the calibrated F1 was slightly lower than zero-shot, and the nested CI crossed zero. We therefore interpret this cell as evidence of a non-random steering effect on the score distribution, not as evidence of robust diagnostic-performance improvement.
Under this stricter interpretation, the NV-Reason disease-axis results remained the most compelling. Pneumonia-text and image-conditioned steering improved calibrated F1 with positive nested CIs, while random-vector controls reinforced the need to interpret the fixed-threshold results cautiously rather than as standalone evidence of diagnostic improvement.

\section{Discussion}
\subsection{Principal Finding}

This benchmark supports a constructive conclusion: inference-time engineering can meaningfully change constrained pneumonia classification operating behavior without fine-tuning, but the correct interpretation depends on calibration and controls. CAA was not a universal diagnostic-performance enhancer in this benchmark, but it was also not inert. It shifted score distributions, altered sensitivity/specificity tradeoffs, and, in the strongest NV-Reason cells, improved calibrated F1 while restoring Yes/No token mass.

The practical message is that activation steering should be framed as operating-point control. That framing is useful for medical AI because many clinical deployments require different sensitivity/specificity targets depending on disease prevalence, workload, and acceptable false-positive rates\cite{Klontzas2026} . The present data support studying CAA and threshold calibration as inference-time controls, which may improve diagnostic capabilities on VLMs.

\subsection{Inference-Time Steering in Context}

CAA was not originally developed for medicine. It emerged from the broader activation-engineering and representation-engineering literature, which treats high-level concepts and behaviors as approximately linear directions in activation space and intervenes on these representations to inspect or influence model behavior\cite{zou2025representationengineeringtopdownapproach, panickssery2024steeringllama2contrastive, turner2024steeringlanguagemodelsactivation, li2024inferencetimeinterventionelicitingtruthful}. Activation Addition introduced inference-time activation modification as a lightweight alternative to prompt engineering or fine-tuning, while Representation Engineering formalized population-level representations as objects that can be monitored and manipulated for transparency and control. CAA extended this logic by computing steering vectors from the mean activation difference between contrastive positive and negative examples and adding those vectors during the forward pass. In Llama 2, CAA showed that such mean-difference vectors can substantially alter model behavior, remain effective when combined with fine-tuning and system prompting, and produce relatively limited degradation of general capabilities\cite{panickssery2024steeringllama2contrastive}.

A useful non-clinical precedent for a measured task-level gain is inference-time intervention(ITI). ITI shifts activations along truthfulness directions in selected attention heads and substantially improved LLaMA performance on TruthfulQA, increasing truthfulness in an instruction-tuned model from 32.5\% to 65.1\%\cite{li2024inferencetimeinterventionelicitingtruthful}. That benchmark result is notable, but the authors highlighted two caveats that motivate the present study: a tradeoff between truthfulness and helpfulness as intervention strength changes, and uncertainty regarding generalization beyond the benchmark. Our study asks the analogous question in a clinical setting: whether an apparent steering gain survives threshold calibration and semantic control conditions.

Subsequent work outside medicine has tempered some of the early optimism around steering vectors in a way that directly anticipates our random-vector findings. In a systematic evaluation across 40 behavioral datasets, Tan et al. found that steerability varies substantially across inputs, that spurious token and position biases can strongly influence apparent steering effectiveness, and that many behaviors remain weakly steerable despite layer and strength sweeps\cite{tan2025analyzinggeneralizationreliabilitysteering}. In several datasets, steering moved predictions in the direction opposite to the intended behavior for a substantial fraction of inputs. Our observation that selected steering vectors for CheXOne and several MedGemma conditions fell within or near the F1 distribution of 20 random unit vectors is a radiology-specific example of the same warning: a vector labeled “pneumonia” is not guaranteed to encode pneumonia, and a large fixed-threshold shift is not, on its own, evidence of disease-specific control.

The contrast with refusal behavior is informative. Prior work in chat language models suggests that refusal of harmful requests is associated with a dominant low-dimensional direction, such that adding or removing this direction can induce or suppress refusal behavior with relatively limited effects on other capabilities \cite{arditi2024refusallanguagemodelsmediated}. In our experiments, the answer-bias axis appears more similar to such a directly accessible response-channel direction, because it perturbs the Yes/No decision tendency itself. By contrast, the disease axes appear partly entangled with this response channel, suggesting that part of the observed steering effect reflects operating-point movement rather than purely disease-specific control.

Within VLMs, inference-time steering is an active and recent line of work. Activation Steering Decoding has been used to suppress hallucination in general large VLMs without retraining by identifying hallucination-related activation patterns and applying bidirectional hidden-state intervention\cite{su-etal-2025-activation}. In radiology report generation, Semantically Decoupled Latent Steering has been proposed to suppress prior-comparison hallucinations by steering hidden states away from hallucination-prone directions while using semantic decomposition and QR-based orthogonalization to reduce entanglement with clinically relevant representations\cite{li2026suppressingpriorcomparisonhallucinationsradiology}. In our study, the observation that an answer-bias vector reproduced disease-axis F1 gains suggests that the disease steering directions were at least partially entangled with the Yes/No response channel rather than representing purely disease-specific control.

\subsection{Why Fixed-Threshold F1 Was Not Enough}
Binary medical AI metrics are threshold-dependent\cite{Klontzas2026}. A score distribution can shift upward without improving the ordering of diseased versus normal cases. If the baseline threshold is poorly matched to the score scale, any intervention that changes calibration can appear to improve F1. MedGemma was the cleanest example: fixed-threshold F1 rose sharply after steering, but calibrated zero-shot F1 was already high, and the nested confidence intervals for steered-minus-zero F1 crossed zero.

The answer-bias axis strengthened this interpretation. A vector built to push affirmative versus negative answers produced gains comparable with disease axes. It shows that activation addition can directly alter the answer channel \cite{arditi2024refusallanguagemodelsmediated}, and that a disease-like fixed threshold gain can be partly or fully explained by response-prior movement.

\subsection{The NV-Reason Exception}
NV-Reason-CXR-3B was the most encouraging result and the only model where disease-axis steering produced calibrated F1 gains with positive nested CIs. The result was also mechanistically more coherent than a threshold-only shift because pneumonia-text and image-conditioned steering increased full-vocabulary Yes/No mass.

The NV-Reason result should not be interpreted as isolated evidence that steering improved diagnostic classification. NV-Reason had very low zero-shot full-vocabulary mass on the Yes/No answer tokens, indicating that the binary prompt did not naturally place the model in a constrained answer mode before steering. Steering increased calibrated F1 while also increasing Yes/No mass by orders of magnitude. Thus, the positive signal likely reflects a combination of answer-channel restoration and disease-related score separation\cite{wiegreffe-etal-2023-increasing}

\subsection{Implications for Future Work}
Future medical activation-steering studies should avoid reporting only fixed-threshold gains. At minimum, credible benchmarks should include calibrated and fixed thresholds, disease and answer-bias axes, reverse-vector and random-vector controls, token-mass diagnostics, ROC-AUC/PR-AUC as rank metrics where appropriate, and per-class sensitivity/specificity. Free-response evaluation is also necessary because clinically relevant VLM output is often a report or explanation, not a single constrained token.

CheXpert\cite{irvin2019chexpertlargechestradiograph} or MIMIC-CXR \cite{johnson2019mimiccxrjpglargepubliclyavailable} would be natural external extensions because they are more widely used than the Kermany pediatric pneumonia split and allow more realistic prevalence and multi-label evaluation. In that setting, Kermany could become a held-out stress test rather than the central benchmark.

\subsection{Limitations}
This study used one public dataset and no external validation cohort. The test split had high pneumonia prevalence, so F1 and precision may not generalize to lower-prevalence settings. Patient demographics and acquisition metadata were unavailable in the local public mirror, preventing subgroup analysis. Benchmark contamination cannot be excluded\cite{sainz-etal-2023-nlp}. No radiologist reader study was performed. Forced-choice scoring evaluates constrained Yes/No logits and does not measure natural-language report quality . Because NV-Reason-CXR-3B appeared poorly aligned to the first-token Yes/No response format at zero-shot, alternative prompt templates, few-shot examples, or constrained decoding could potentially raise baseline Yes/No mass and reduce the apparent benefit of steering. We therefore interpret the NV-Reason finding as a constrained-scoring sensitivity result rather than a standalone diagnostic-performance improvement.  Random-vector controls show that activation injection itself can produce large model-dependent shifts, so steering effects require careful control comparisons.

\section{Conclusion}
Inference-time engineering changed pneumonia-classification operating behavior across three CXR VLMs without fine-tuning. The strongest positive signal was NV-Reason disease-axis steering, where calibrated F1 improved under conservative nested thresholding. The strongest cautionary signals were MedGemma answer-bias equivalence and CheXOne threshold sensitivity. CAA and threshold calibration should therefore be viewed as promising tools for probing and controlling medical VLM operating points, not yet as validated methods for improving clinical diagnosis.

\section*{Data and Code Availability}

The study was conducted using the publicly available Chest X-Ray Pneumonia dataset hosted on Hugging Face. All code required to reproduce the benchmark, including environment specifications, experiment scripts, and auditing utilities, is publicly available at \url{https://github.com/eduardofarina/repe_benchmarking}. Generated outputs and intermediate artifacts are not version-controlled and can be reproduced by executing the documented workflows provided in the repository.

\newpage

\bibliographystyle{unsrt}  
\bibliography{references}

\end{document}

%% file: table1.tex
\begin{table}[htbp]
\centering
\caption{Steering axes evaluated in the pneumonia CAA benchmark.}
\label{tab:steering_axes}
\small
\begin{tabularx}{\textwidth}{lXXX}
\toprule
\textbf{Axis} & \textbf{Positive side} & \textbf{Negative side} & \textbf{Purpose} \\
\midrule
Pneumonia text &
30 pneumonia-positive text statements &
30 pneumonia-negative text statements &
Disease-language steering \\

Answer bias &
30 affirmative Yes-style prompts &
30 negative No-style prompts &
Direct control for answer-prior steering \\

Image-conditioned pneumonia &
30 pneumonia development images with the classification question &
30 normal development images with the same question &
Image-grounded steering with identical text \\
\bottomrule
\end{tabularx}
\end{table}

%% file: table2.tex
\begin{table*}[htbp]
\centering
\caption{Expanded-scale calibrated performance and control summary. CI = confidence interval; PR-AUC = precision-recall area under the curve; mass = mean $P_{\mathrm{vocab}}(\mathrm{Yes}) + P_{\mathrm{vocab}}(\mathrm{No})$.}
\label{tab:expanded_performance}
\scriptsize
\begin{tabularx}{\textwidth}{llccccccX}
\toprule
\textbf{Model} & \textbf{Axis} & \textbf{Zero F1} & \textbf{Steered F1} & \textbf{Sens/Spec} & \textbf{PR-AUC} & \textbf{Mass} & \textbf{Nested 95\% CI} & \textbf{Interpretation} \\
\midrule
CheXOne & Answer bias & 0.853 & 0.807 & 0.938/0.355 & 0.896 & 9.99e-01 & -0.067 to 0.003 & Not robust \\
CheXOne & Image-conditioned & 0.853 & 0.852 & 0.931/0.577 & 0.850 & 1.00e+00 & -0.052 to 0.040 & Not robust \\
CheXOne & Pneumonia text & 0.853 & 0.867 & 0.941/0.615 & 0.875 & 1.00e+00 & -0.046 to 0.063 & Not robust \\
MedGemma & Answer bias & 0.891 & 0.887 & 0.913/0.756 & 0.942 & 7.25e-01 & -0.051 to 0.032 & Not robust \\
MedGemma & Image-conditioned & 0.891 & 0.899 & 0.923/0.782 & 0.946 & 5.83e-01 & -0.026 to 0.039 & Not robust \\
MedGemma & Pneumonia text & 0.891 & 0.885 & 0.944/0.684 & 0.950 & 5.14e-01 & -0.018 to 0.036 & Not robust \\
NV-Reason & Answer bias & 0.769 & 0.798 & 0.995/0.167 & 0.886 & 2.02e-04 & 0.018 to 0.091 & Response-channel control \\
NV-Reason & Image-conditioned & 0.769 & 0.873 & 0.967/0.585 & 0.864 & 2.46e-01 & 0.063 to 0.124 & Positive calibrated signal \\
NV-Reason & Pneumonia text & 0.769 & 0.862 & 0.936/0.607 & 0.916 & 5.27e-02 & 0.039 to 0.119 & Positive calibrated signal \\
\bottomrule
\end{tabularx}
\end{table*}